\documentclass{article}

\usepackage{arxiv}

\usepackage{graphicx}
\usepackage{algorithm}
\usepackage{algorithmic}
\usepackage{longtable}
\usepackage{subfigure}
\usepackage{amsmath}
\usepackage{booktabs}
\usepackage{supertabular}
\usepackage{multicol}
\usepackage{multirow}

\title{CODE-AE: A Coherent De-confounding Autoencoder for Predicting Patient-Specific Drug Response From Cell Line Transcriptomics}

\author{
 Di He \\
  PhD program in Computer Science \\
  Graduate Center, City University of New York \\
  New York City, 10016, USA\\
  \texttt{dhe@gradcenter.cuny.edu} \\
   \And
 Lei Xie\thanks{This author is also affiliated with Feil Family Brain \& Mind Research Institute, Weill Cornell Medicine, Cornell University} \\
  Department of Computer Science \\ 
  Hunter College, City University of New York \\ 
  New York City, 10065, USA \\
  \texttt{lxie@iscb.org} \\
}

\begin{document}

\maketitle

\begin{abstract}
Accurate and robust prediction of patient's response to drug treatments is critical for developing precision medicine. However, it is often difficult to obtain a sufficient amount of coherent drug response data from patients directly for training a generalized machine learning model. Although the utilization of rich cell line data provides an alternative solution, it is challenging to transfer the knowledge obtained from cell lines to patients due to various confounding factors. Few existing transfer learning methods can reliably disentangle common intrinsic biological signals from confounding factors in the cell line and patient data. In this paper, we develop a Coherent Deconfounding Autoencoder (CODE-AE) that can extract both common biological signals shared by incoherent samples and private representations unique to each data set, transfer knowledge learned from cell line data to tissue data, and separate confounding factors from them. Extensive studies on multiple data sets demonstrate that CODE-AE significantly improves the accuracy and robustness over state-of-the-art methods in both predicting patient drug response and de-confounding biological signals. Thus, CODE-AE provides a useful framework to take advantage of \textit{in vitro} omics data for developing generalized patient predictive models. The source code is available at https://github.com/XieResearchGroup/CODE-AE.\\
\\
\textbf{Contact:} lxie@iscb.org\\
\end{abstract}
\section{Introduction}
Transcriptomics is a powerful technique to characterize cellular activity under various conditions, allowing researchers to uncover the underlying associations among genes, biological pathways, diseases, and environmental factors. Hence, this data source has been widely explored by studies ranging from regulatory gene identification \cite{danaee2017deep, gtex2017genetic} to disease biomarker discovery \cite{goossens2015cancer}. In particular, it has been utilized to construct predictive machine learning models for drug response, such as in \cite{CCLE, sakellaropoulos2019deep}. However, the success of such predictive models largely relies on the availability of sufficient amounts of data with coherent and comprehensive annotations. In clinical, we are often short of a large number of patient samples with drug treatment and response history. For this reason, most drug response predictive studies to date have mainly utilized transcriptomic profiles from panels of in-vitro cancer cell lines as input features. Although such an approach is promising, the utility of drug response models built with in-vitro data is often limited when applied to real patients due to the genetic and environmental differences between in-vitro cell lines and patient-derived tissue samples and confounding factors that may mask intrinsic biological signals. 

To address the above challenges, we propose a Coherent Cell line-Tissue Deconfounding Autoencoder (CODE-AE) that can extract both common biological signals shared by incoherent samples and private representations unique to them, transfer knowledge learned from cell line data to tissue data, and separate confounding factors from them. CODE-AE will allow us to generalize existing cell line omics data for robust predictive modeling of drug response to new patients, a critical component for patient-specific drug screening and precision medicine. Specifically, in CODE-AE, we devise a self-supervised training scheme to construct an encoding module that can be easily tuned to adapt to the different downstream tasks. For the self-supervised training of encoder, we leverage both unlabeled cell line and tissue samples.

As a demonstration of the potential of CODE-AE in precision medicine, we apply CODE-AE to predicting chemotherapy resistance for patients, which is a significant obstacle to effective cancer therapy. Lack of effective personalized chemotherapy tailored to individual patients often leads to unnecessary suffering and reduces the chances of patient's overall survival. Our extensive studies show that CODE-AE significantly outperforms state-of-the-art methods in terms of both accuracy and robustness. Thus CODE-AE provides a useful framework to take advantage of rich \textit{in vitro} omics data for developing generalized patient predictive models. 

\section{Related Work}
Our goal is to learn an encoding function that maps a gene expression profile to a low-dimensional vector (embedding) dominated by intrinsic biological signals from both in-vitro cell lines and in-vivo tissue samples. For such representation learning problems, a well-established technique is an autoencoder and its variants due to their relatively low reliance on labeled data availability. An autoencoder \cite{autoencoders} is a neural network trained to learn how to copy its input to output. Typically it has one bottleneck hidden layer that learns to capture the hidden patterns underlying the raw input. Denoising autoencoder (DAE) \cite{dae} is an extension to the standard autoencoder, which attempts to minimize the reconstruction cost with the input corrupted by some form of noise. The goal of denoising autoencoders is to make the reconstruction function resist small but finite-sized perturbations of the input. In other words, the encoder and decoder are forced to learn the structure of the distribution of input data implicitly. Variational autoencoders (VAE) \cite{vae} are another regularized version of autoencoder, which gained popularity due to its mathematical elegance and capability to enable the generative process. Both DAE and VAE have been widely adapted to characterize gene expression profiles \cite{ching2018opportunities}.

Another area of studies within our consideration is domain adaptation techniques, particularly feature-based domain adaptation methods \cite{weiss2016survey}, which aim to learn a domain-variant feature representation by minimizing the discrepancy across different domains. Methods such as deep domain confusion \cite{DCC2014}, Deep Coral \cite{deepcoral2016} focus on utilizing statistical distribution discrepancy metrics to achieve the desired domain discrepancy reduction. By adopting the popular adversarial training scheme, Domain adversarial neural network (DANN) \cite{dann2016} learned domain invariant features by a minimax game between the domain classifier and the feature generator with layer sharing and customized gradient reversal layer. Similarly, adversarial discriminative domain adaptation (ADDA) \cite{adda2017} achieved domain adaptation via a general framework consisting of discriminative modeling, untied weight sharing, and a GAN loss. In addition to the discrepancy reduction approaches mentioned above, encoder-decoder models are also widely used in domain adaptation, where domain invariant features are learned via shared intermediate representation. In contrast, domain-specific features are preserved with reconstruction loss. Representative works include marginalized denoising autoencoder \cite{mDAE}, multi-task autoencoders \cite{mtAE}, deep reconstruction classification network \cite{drcn}. Domain separation network (DSN) \cite{dsn} was proposed to separate private representations for each domain explicitly and shared representations across domains. The shared representation is learned similarly as DANN \cite{dann2016} or with MMD \cite{MMD}, while the private representations are learned via orthogonality constraint against the shared representation. DSN can achieve better generalization across domains with the reconstruction through the concatenation of shared and private representations than other methods. More recently, adversarial de-confounding autoencoder (ADAE) \cite{adae} combined the idea of the encoder-decoder model with adversarial training as in DANN \cite{dann2016} to deconfound the unwanted factors prevalent in biological data sets to extract salient biology features.

\section{Methods}
\subsection{Our Approach: COherent DE-confounding AutoEncoder (CODE-AE)}
We proposed the CODE-AE to generate biologically informative gene expression embeddings applicable to transfer between in-vitro and patient samples. CODE-AE employed the standard auto-encoder as the backbone to leverage the unlabeled gene expression data sets. Inspired by the work on factorized latent space \cite{fols} and domain separation network \cite{dsn}, we encoded the samples (from cell lines or tumor tissues) into two non-redundant embeddings, namely private embeddings and shared embeddings. The first one is designed to capture the cell line or tissue-specific information. The latter contains the deconfounded biological meaningful information used to transfer knowledge across cell lines and tissues. 
\subsubsection{CODE-AE Base}
\begin{figure*}[!t]
\centering
\includegraphics[width=\textwidth]{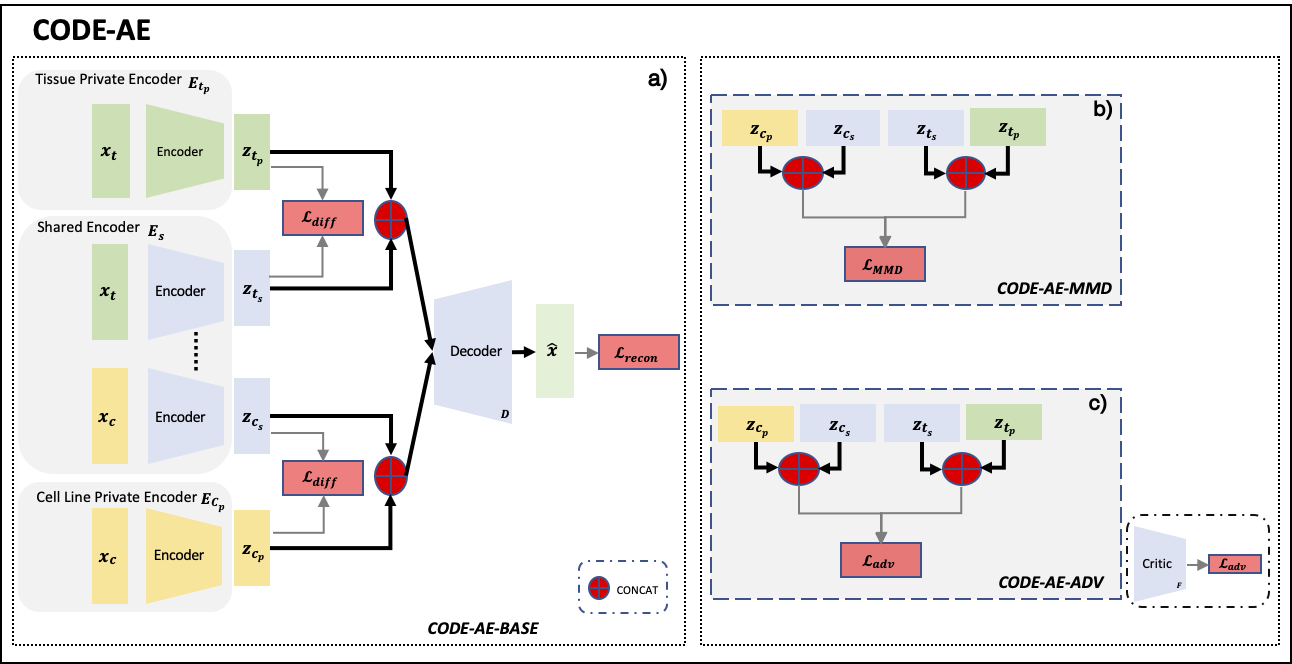}
\caption{\textbf{COherent Deconfounding AutoEncoder (CODE-AE) framework}. a) CODE-AE Base architecture: A layer-tying shared encoder \(\textbf{E}_s\) learns to map both cell line and tissue samples to deconfounded biological meaningful embeddings. Private encoders \(\textbf{E}_{\cdot _ {p}}\) learn to represent cell line/tissue specific information as private embeddings. A shared decoder \(\textbf{D}\) reconstructs the input samples through the concatenation of private and shared embeddings and the reconstruction quality is measured with \(\mathcal{L}_{recon}\). The private and shared embeddings are pushed apart through soft subspace orthogonality loss \(\mathcal{L}_{diff}\) b) Appended module of CODE-AE-MMD: the concatenation of private and shared embeddings are kept similar with \(\mathcal{L}_{MMD}\) c) Appended module of CODE-AE-ADV: the concatenation of private and shared embeddings are kept similar with \(\mathcal{L}_{adv}\), where \(\mathcal{L}_{adv}\) is in the form of min-max optimization between critic network \(\textbf{F}\) and other components.}
\label{Fig:01}
\end{figure*}
As shown in Figure~1\vphantom{\ref{Fig:01}}, the CODE-AE takes expression vectors from in-vitro cell lines and patient tumor tissue samples as input. 
\linespread{1} Let \(\emph{X}_t = \left \{ x_t^{(i)} \right \}_{i=1}^{N_t}\) and \(\emph{X}_c = \left \{ x_c^{(i)} \right \}_{i=1}^{N_c}\) represent the unlabeled data set of \(N_t\) patient tumor tissue samples and \(N_c\) in-vitro cancer cell line samples, respectively. Each sample \(x\) will be encoded into two separate embeddings through its corresponding cell line or tissue private encoder \(\textbf{E}_{\cdot _ {p}}\) and also the weight-sharing encoder \(\textbf{E}_{s}\). The concatenation of these two embeddings of each sample is expected to be able to reconstruct the original gene expression vector \(x\) through a shared decoder \(\textbf{D}\), and the reconstruction is done as,
\begin{equation}
    \widehat{\textbf{x}_{\cdot}^{(i)}} = D(\textbf{E}_{s}(\textbf{x}_{\cdot}^{(i)}) \bigoplus \textbf{E}_{\cdot _ {p}}(\textbf{x}_{\cdot}^{(i)}))
\end{equation}
where \(\textbf{x}_{\cdot}^{(i)}\) represents the input gene expression profile, \(\widehat{\textbf{x}_{\cdot}^{(i)}}\) is the corresponding reconstructed input sample through the autoencoder component. \(\bigoplus\) stands for the vector concatenation operation.
We measure the quality of autoencoder reconstruction through the mean squared error between the original samples and the reconstruction output as below, 
\begin{equation}
    \mathcal{L}_{recon} = \frac{1}{N_c}\sum_{i=1}^{N_{c}}\left \| \textbf{x}_{c}^{(i)}-\widehat{\textbf{x}_{c}^{(i)}} \right \|_2^2 + \frac{1}{N_t}\sum_{i=1}^{N_{t}}\left \| \textbf{x}_{t}^{(i)}-\widehat{\textbf{x}_{t}^{(i)}} \right \|_2^2
\end{equation}
In our formulation, we factorized each sample's latent space into two different subspaces to capture both domain specific and common information separately. To minimize the redundancy between the factorized latent spaces, we included an additional penalty term, \(\mathcal{L}_{diff}\) in the form of orthogonality constraint.
The difference loss \(\mathcal{L}_{diff}\), is applied to both cell line and tissue samples and encourages the shared and private encoder to encode different aspects of the inputs. We define the loss via soft subspace orthogonality constraint as below, 
\begin{equation}
    \mathcal{L}_{diff} = \left \| \textbf{Z}_{c_{s}}^{T} \textbf{Z}_{c_{p}}\right \|_F^2 + \left \| \textbf{Z}_{t_{s}}^{T} \textbf{Z}_{t_{p}}\right \|_F^2
\end{equation}
where \(\textbf{Z}_{\cdot_{s}}\) are the embedding matrices whose rows are the shared embedding for cell line or tissue samples, while \(\textbf{Z}_{\cdot_{p}}\) are the embedding matrices whose rows are the private embedding for cell line or tissue samples. It is obvious that \(\mathcal{L}_{diff}\) tends to push the embeddings to meaningless all-zero-valued vectors. To avoid such scenario, we append an additional instance normalization layer after the output layer of each encoder to avoid embeddings with minimal norm. Lastly, the loss for CODE-AE-BASE is defined with the weighted combination between \(\mathcal{L}_{recon}\) and \(\mathcal{L}_{diff}\) as below, 
\begin{equation}
    \mathcal{L}_{code-ae-base} =  \mathcal{L}_{recon} + \alpha \mathcal{L}_{diff}
\end{equation}
where \(\alpha\) is the embedding difference loss coefficient.
\subsubsection{CODE-AE Variants}
With CODE-AE-BASE, we could split cell line or tissue sample's inherent information into the private and shared streams. However, in our baseline experiments, we often found that it was sub-optimal or demonstrated varied performance. Thus, in this section, we proposed two variants that showed better and generally more stable performance. Under the CODE-AE framework, for each input sample, CODE-AE factorized it into two embeddings. The concatenation of these two embeddings is considered as the new representation of the original input. Given that all samples in our consideration are gene expression profiles regardless of cell line or patient, we assumed that the new representation of original input in the factorized latent space close to each in terms of distributional differences. 

\textbf{CODE-AE-MMD}. The first variant, named CODE-AE-MMD, utilized the well known maximum mean discrepancy \cite{MMD} as the distance measurement between the latent representation of cell line and tissue samples. Maximum Mean Discrepancy (MMD) loss \cite{MMD} is a kernel-based distance function between samples from two distributions. In particular, we used an approximate version of exact MMD loss in CODE-AE-MMD as below,
\begin{align}
\mathcal{L}_{MMD}(\textbf{Z}_{c}, \textbf{Z}_{t}) & = \frac{1}{N_2}\sum_{i, j = 0}^{N}\kappa (\textbf{{z}}_{c}^{(i)}, \textbf{{z}}_{c}^{(j)}) + \frac{1}{N_2}\sum_{i, j = 0}^{N}\kappa (\textbf{{z}}_{t}^{(i)}, \textbf{{z}}_{t}^{(j)}) \nonumber\\
& - \frac{2}{N^2}\sum_{i, j = 0}^{N}\kappa (\textbf{{z}}_{c}^{(i)}, \textbf{{z}}_{t}^{(j)}) 
\end{align}
where \(\textbf{Z}_{c}, \textbf{Z}_{t}\) are embedding matrices for cell line and tissue samples respectively, whose rows are the concatenations of each sample's private and shared embedding. \(\textbf{{z}}_{\cdot}^{(i)}, \textbf{{z}}_{\cdot}^{(j)}\) are the \(i\)-th or \(j\)-th samples' corresponding embedding vectors. In practice, \(N\) will be the batch size. Accordingly, the loss of CODE-AE-MMD is given as below, 
\begin{equation}
    \mathcal{L}_{code-ae-mmd} =  \mathcal{L}_{code-ae-base} + \beta \mathcal{L}_{MMD}
\end{equation}
where \(\beta\) is the MMD loss coefficient.

\textbf{CODE-AE-ADV}
The second variant, CODE-AE-ADV, employed adversarial training to push the representations of cell line and tissue samples to be similar to each other. Specifically, we appended a critic network \(F\) that scores representations with the objective that consistently gives higher scores for representations of cancer cell line samples. The encoders for tissue samples are given an additional objective to generate the embedding that could fool the critic network to produce high scores.
In this manner, critic network and tissue sample encoders will play a min-max game in the form of an alternative training schedule, which is adopted by Wasserstein generative adversarial networks \cite{wgan}. To avoid unstable training commonly existing in alternative training schedules, instead of standard WGAN \cite{wgan} we used the WGAN with gradient penalty \cite{wgangp}. Its affiliated loss terms are defined as below,
\begin{equation}
    \mathcal{L}_{adv}: \left\{\begin{matrix}
\mathcal{L}_{critic} = \frac{1}{N_t}\sum_{i=1}^{N_t}F(\textbf{z}_t^{(i)})-\frac{1}{N_c}\sum_{i=1}^{N_c}F(\textbf{z}_c^{(i)}) 
\\ + \lambda (\left \| \bigtriangledown_{\widetilde{\textbf{z}}}F(\widetilde{\textbf{z}}) \right \|_2-1)^2 
\\
\\
\mathcal{L}_{gen} = -\frac{1}{N_t}\sum_{i=1}^{N_t}F(\textbf{z}_t^{(i)})
\end{matrix}\right.
\end{equation}
where \(\textbf{z}_{\cdot} = \textbf{z}_{\cdot_{s}} \bigoplus \textbf{z}_{\cdot_{p}}\) stands for new representation of input and \(\widetilde{\textbf{z}} = \epsilon \textbf{z}_c + (1-\epsilon)\textbf{z}_t \) and \(\epsilon \sim \textbf{U}(0,1)\). A detailed CODE-AE-ADV learning procedure can be found in (Procedure \ref{alg1}).
\algsetup{indent=2em}
\floatname{algorithm}{Procedure}
\begin{algorithm}[H]
\small
\caption{CODE-AE-ADV training}
\label{alg1}
\textbf{Input}: \(\{\textbf{x}_{c}^{(i)}\}_{i=1}^{N_c}, \{\textbf{x}_{t}^{(i)}\}_{i=1}^{N_t}\)
\begin{algorithmic}[1]
\REQUIRE $N$, the batch size
\\ $\lambda$, generator loss coefficient
\\ $n_{w}$, number of warm-up epochs
\\ \(n_{t}\), number of training epochs
\\ \(n_{critic}\), number of steps per encoders update
\FOR{epoch $=1$ to $n_{w}$}
\FOR{\(t=1\) to \(\frac{min(N_c, N_t)}{N}\)} 
\STATE sample \(\{\textbf{x}_{c}\}\) of size \(N\) from \(\{\textbf{x}_{c}^{(i)}\}_{i=1}^{N_c}\) (w/o. rep)

\STATE sample \(\{\textbf{x}_{t}\}\) of size \(N\) from \(\{\textbf{x}_{t}^{(i)}\}_{i=1}^{N_t}\) (w/o. rep)
\STATE Update \(E_{t_p}\), \(E_{c_p}\), \(E_{s}\), \(D\) with \(\mathcal{L}_{code-ae-base}\)
\ENDFOR
\ENDFOR
\FOR{epoch $=1$ to $n_{t}$}
\FOR{\(t=1\) to \(\frac{min(N_c, N_t)}{N}\)} 
\STATE sample \(\{\textbf{x}_{c}\}\) of size \(N\) from \(\{\textbf{x}_{c}^{(i)}\}_{i=1}^{N_c}\) (w/o. rep)

\STATE sample \(\{\textbf{x}_{t}\}\) of size \(N\) from \(\{\textbf{x}_{t}^{(i)}\}_{i=1}^{N_t}\) (w/o. rep)
\STATE Update \(F\) with \(\mathcal{L}_{critic}\) 
\IF{t \(\% n_{critic} == 0\) }
\STATE Update \(E_{t_p}\), \(E_{c_p}\), \(E_{s}\), \(D\) with \(\mathcal{L}_{code-ae-base} + \lambda  \mathcal{L}_{gen} \)
\ENDIF
\ENDFOR
\ENDFOR
\end{algorithmic}
\end{algorithm}
After the encoder training with unlabeled data as mentioned above, the shared encoder \(\textbf{E}_s\) could be used to directly generate the deconfounded biological meaningful embedding vectors or append a neural network module for specific downstream tasks. In the latter case, strategies such as gradual unfreezing and decayed learning rate schedule could be adopted to improve task-specific performance further, as shown in our following experiments. 
\section{Experiments Setup}
\subsection{Baseline models}
We compared CODE-AE with the following base-line models: standard autoencoder (AE) \cite{autoencoders}, denoising autoencoder (DAE) \cite{dae}, and variational autoencoder (VAE) \cite{vae} as well as representative domain adaptation methods including deep coral (CORAL) \cite{deepcoral2016} and domain separation network (DSN) \cite{dsn} of both MMD (DSN-MMD) and adversarial (DSN-ADV) training variants. Furthermore, we included a more recent adversarial deconfounding autoencoder (ADAE) \cite{adae} given its similar formation as DANN \cite{dann2016} and state-of-the-art performance in transcriptomics data sets. For fair comparisons, all the encoder and decoder trained in the experiments share the same architecture. Specifically, the hidden representation is of dimension 128. The encoders and decoder are 2-layer neural network modules of dimension (512, 256) and (256, 512), respectively, with the rectified linear activation function. Appended modules such as critic network in CODE-AE-ADV, and classifier network used for fine-tuning are 2-layer neural networks of dimension (64, 32) with rectified linear activation, have one output node with linear activation in critic network and sigmoid activation in classifier networks. Further, the loss weight terms in CODE-AE-MMD and CODE-AE-ADV are all simply specified as 1.0.

For reference, we also included the classification performances of similarly fine-tuned randomized initialized encoder (labeled as MLP). In addition, the elastic net classifier (labeled as EN) trained on original cell line gene expression profiles also included in the comparison.
\subsection{Data sets}
We evaluated the performance of CODE-AE with a practical problem: predict chemotherapy resistance given gene expression profiles of patients while training the predictive model \textit{only} using the gene expression of cancer cell lines. We collected the cancer cell line gene expression profiles from the DepMap portal \cite{CCLE2} and corresponding drug sensitivity data from GDSC \cite{GDSC1, GDSC2}. Additionally, we collected patients' tumor tissue gene expression profiles from the Xena portal \cite{xena}. In total, we gathered 1305 cancer cell lines and 9808 patient tumor tissue samples with corresponding gene expression profiles, respectively. All gene expression data are metricized by the standard transcripts per million base for each gene, with additional log transformation.

Clinical chemotherapy resistance can be defined as either a lack of reduction in the size of tumor following chemotherapy or the occurrence of clinical relapse after an initial “positive response to treatment” \cite{ben2019resistance}. Hence, we extracted data sets to assess these two aspects. The patient clinical drug response was acquired from a recent work \cite{tcga_dataset}, where patients' clinical response records of two chemotherapy agents Gemcitabine and Fluorouracil from The Cancer Genome Atlas (TCGA) \cite{tcga} were extracted. The patients were split into two groups: responders who had a partial or complete response and non-responders who had a progressive clinical disease or stable disease diagnosis. Only patients on single-drug therapy through the entire duration of treatment were retained in the study. 

In addition to using clinical diagnosis to indicate patients' drug responses towards a particular drug, we extracted patients "new tumor events days after treatment" from TCGA \cite{tcga} as the standard to divide patient into responders and non-responders. The median number of days of new tumor events was used as the threshold. Similar to the above data set from \cite{tcga_dataset}, we only included patients on single-drug therapy through the entire treatment duration in this test data set. Due to the limited size of patients' samples, we only included patients who received Cisplatin and Temozolomide in this test data set. 
\subsection{Training procedure}
We selected the top 1000 varied genes measured by the percentage of unique values for cancer cell lines and tumor tissue samples separately. Then we combined the two sets of top 1000 varied genes as the input features. The union has 1424 unique genes in total. We first pre-train different variants of the autoencoders as mentioned above using the same unlabeled samples from both cancer cell lines and tumor tissues. Then we fine-tune the pre-trained encoders with appended classification module over labeled cancer cell line samples and corresponding drug sensitivity data. Specifically, we first selected all cell lines with corresponding drug sensitivity measured in the area under the drug response curve (AUC). We further categorized these cancer cell lines' sensitivity against this drug into binary labels, namely resistant or responsive. In particular, for the drug sensitivity measured in AUC, it was presented as the fraction of the total area under drug response curve between the highest and lowest screening concentration in GDSC \cite{GDSC1, GDSC2}. We set the AUC threshold as the value that produced the best classification performance of an elastic net classifier given different drugs. 
The number of training cell line samples (responsive/resistant) and test tumor tissue samples for different drugs sensitivity prediction tasks are summarized in Table~1\vphantom{\ref{Tab:01}}.
\begin{table}[h!]
\centering 
\caption{Training/Test samples class distribution}  
\label{Tab:01}
\begin{tabular}{ccccc}
\hline
 & \multicolumn{2}{c}{Training (cancer cell lines)} & \multicolumn{2}{c}{Test (patient tumor tissues)} \\ \hline
 & responsive & resistant & responsive & resistant \\ \hline
Gemcitabine & 301 & 376 & 37 & 55 \\
Fluorouracil & 23 & 644 & 34 & 24 \\
Cisplatin & 291 & 377 & 20 & 20 \\
Temozolomide & 19 & 660 & 25 & 24 \\ \hline
\end{tabular}
\end{table}
During the fine-tuning stage, the cell line samples were split into ten stratified folds (according to cancer types). In one evaluation iteration, nine out of ten folds of the samples were used as the training set. The remaining one fold of samples was used as the validation data set for early stopping of fine-tuning process.
\subsection{Performance evaluation}
We choose the area under the receiver operating curve (AUROC) as the measurement metric due to their insensitivity to changes in the test data set's class distribution \cite{roc}. The model performance was measured in AUROC over the patient tissue expression data and corresponding drug response records. The performance of different methods was compared by the average of AUROCs of ten iterations. It is noted that only cell line data were used for the model training and hyperparameter selections, and all patient data were purely used for the testing. 
\section{Results and Discussion}
\subsection{Chemotherapy resistance prediction}
\begin{figure*}[!t]
    \includegraphics[width=1.0\linewidth]{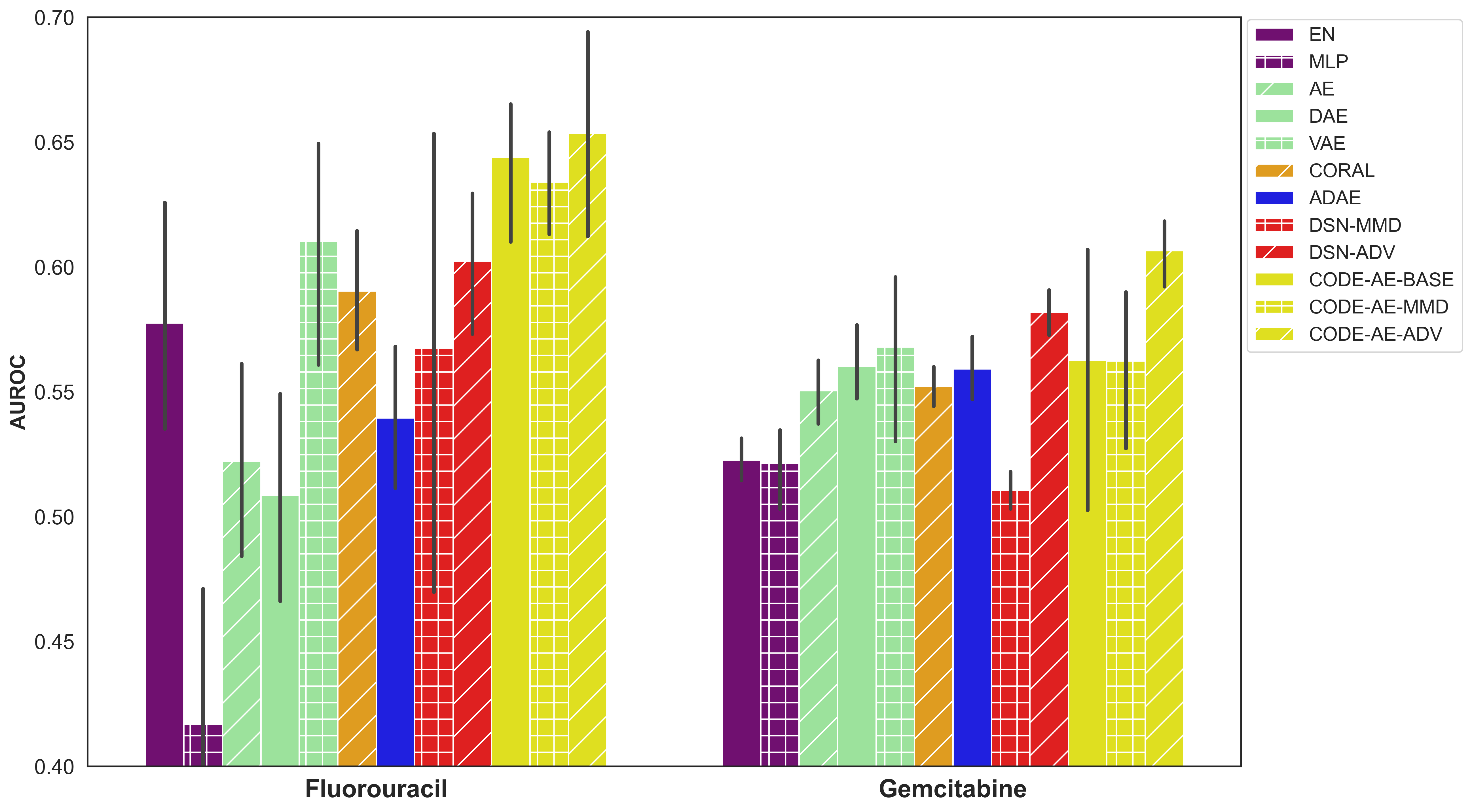}
\caption{The left and right figures show the performance of different methods trained with the drug sensitivity of cancer cell lines for predicting patients' response to chemotherapy agents Gemcitabine and Fluorouracil, respectively. Patients' drug responses are measured with clinical diagnosis.}
\end{figure*}
\begin{figure*}[!t]
\centerline{\includegraphics[width=1.0\linewidth]{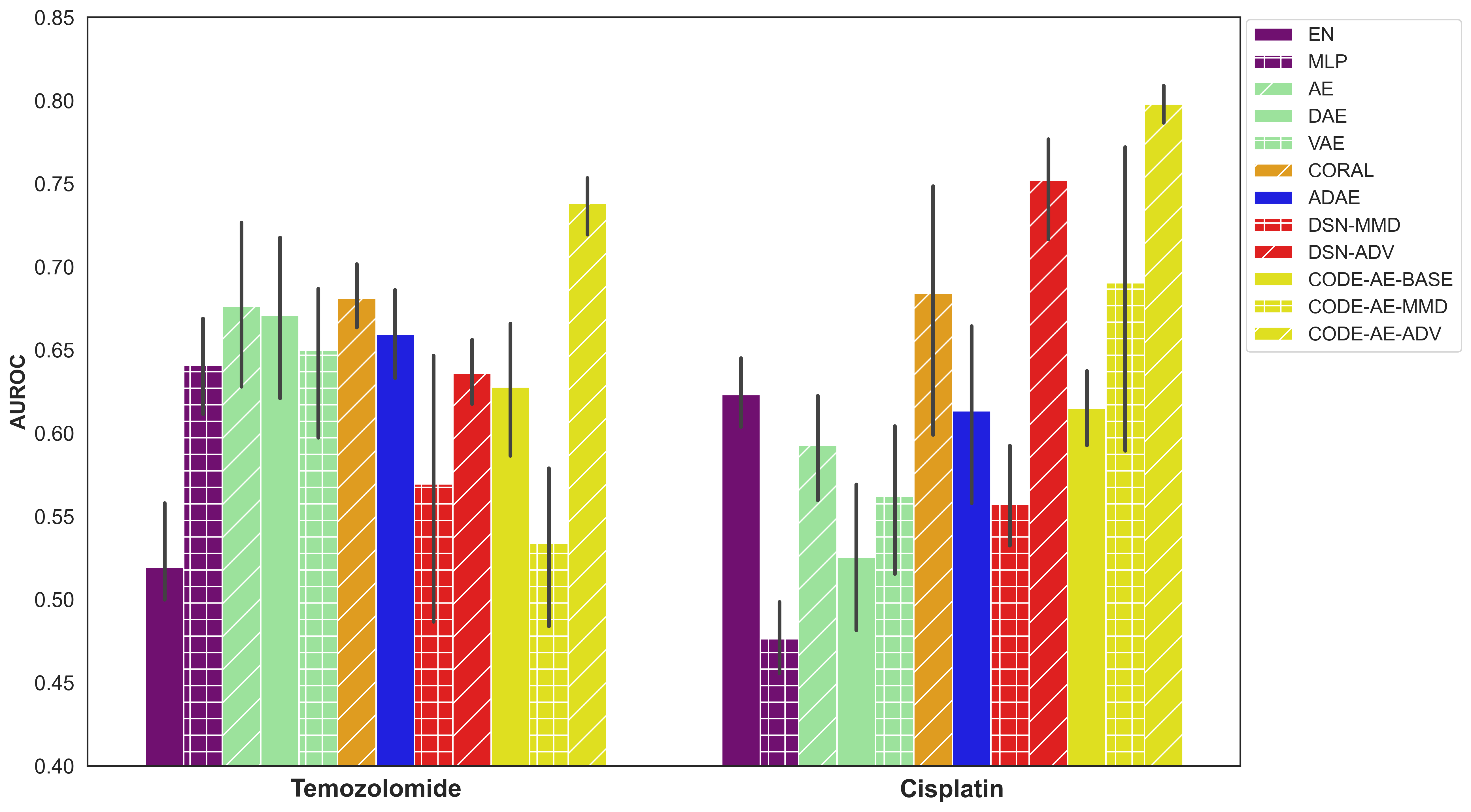}}
\caption{The left and right figures show the performance of different methods trained with the drug sensitivity of cancer cell lines for for predicting patients' response to chemotherapy agent Cisplatin and Temozolomide, respectively. Patients' drug responses are categorized based on number of days of new tumor events after treatment.}\label{fig:03}
\end{figure*}
We evaluated clinical chemotherapy resistance in two aspects: either a lack of reduction in the size of tumor following chemotherapy or the occurrence of clinical relapse after an initial “positive response to treatment” \cite{ben2019resistance} as described in Section 4.2. The results are shown in Fig~2\vphantom{\ref{fig:02}} and Figure~3\vphantom{\ref{fig:03}}, respectively. Based on these results, we have the following observations.
\begin{table}[h!]
\centering 
\caption{Average AUROC performance t-tests p-values }
\label{Tab:02} 
\begin{tabular}{ccc}
\hline
\textbf{Drug} & \textbf{Best Non CODE-AE method} & \textbf{P-value} \\ \hline
\textbf{Gemcitabine} & DSN-ADV & 0.010894 \\
\textbf{Fluorouracil} & VAE & 0.205175 \\
\textbf{Cisplatin} & DSN-ADV & 0.033298 \\
\textbf{Temozolomide} & CORAL & 0.000557 \\ \hline
\end{tabular}
\end{table}

\textbf{Observation 1. CODE-AE-ADV significantly outperforms the state-of-the-art methods.} On average, CODE-AE-ADV shows the highest value of AUROC for all four drugs tested. We performed a two-sample t-test on the AUROC performance between CODE-AE-ADV and the best non-CODE-AE method for each drug, as shown in Table~2\vphantom{\ref{Tab:02}}. Among three of four drugs (Gemcitabine, Cisplatin, and Temozolomide), CODE-AE-ADV significantly outperforms other methods. For the drug Fluorouracil, the performance CODE-AE-ADV does not significantly improve the second-best performer VAE, but is significantly better than other state-of-the-art methods CORAL and ADAE. Moreover, CODE-AE-ADV is more robust than other methods, as demonstrated by relatively smaller variants of the performance.   

\textbf{Observation 2. Private and shared embeddings of cell line and patient data contribute to improving the performance of transfer learning.} The major difference between CODE-AE-ADV and ADAE is to disentangle shared and private embeddings between cell lines and patient tissues. It is also true for the difference between DSN-ADV and ADAE. For all four drugs, CODE-AE-ADV significantly outperforms ADAE. Among three of four drugs (Fluorouracil, Gemcitabine, and Cisplatin), DSN-ADV also performs significantly better than ADAE. For the drug Temozolomide, the performance difference between DSN-ADV and ADAE is not statistically significant (p-value = 0.194747). 

\textbf{Observation 3. Adversarial loss outperforms MMD loss.} For all four drugs, CODE-AE-ADV and DSN-ADV significantly outperform their variants CODE-AE-MMD and DSN-MMD, respectively. The only difference between them is the loss function, adversarial loss or MMD loss. Clearly, transfer learning between cell lines and patients may benefit from the adversarial training that provides an effective way to sample the learning space. 
\subsection{Transferability of deconfounded representation of CODE-AE}
\begin{table}[!h]
\centering
\caption{Performance comparison on cancer subtype prediction. The best and the second best performances are highlighted and underlined, respectively.} 
\label{Tab:03} 
\begin{tabular}{ccccc}
\hline
\textbf{} & \multicolumn{2}{c}{\textbf{Female-\textgreater{}Male}} & \multicolumn{2}{c}{\textbf{Male-\textgreater{}Female}} \\
 & AUROC & AUPRC & AUROC & AUPRC \\ \hline
ADAE & \underline{0.9038$\pm$0.0081} & \underline{0.9621$\pm$0.008} & 0.9264$\pm$0.0043  & 0.9755$\pm$0.0018 \\
CORAL & 0.8793$\pm$0.0208  & 0.9432$\pm$0.0042  & \textbf{0.9444$\pm$0.0057} & \textbf{0.9862$\pm$0.002} \\
DSN-MMD & 0.6772$\pm$0.0187 & 0.8715$\pm$0.026 & 0.6693$\pm$0.0584 & 0.9145$\pm$0.0189 \\
DSN-ADV & 0.7513$\pm$0.0341 & 0.9016$\pm$0.0086 & 0.8518$\pm$0.035 & 0.9608$\pm$0.0098 \\
CODE-AE-BASE & 0.603$\pm$0.0437 & 0.8613$\pm$0.0001 & 0.6276$\pm$0.0687 & 0.9112$\pm$0.0079 \\
CODE-AE-MMD & 0.9221$\pm$0.0186 & 0.9683$\pm$0.0056 & 0.939$\pm$0.0039 & 0.9826$\pm$0.0021 \\
CODE-AE-ADV & \textbf{0.9319$\pm$0.0018} & \textbf{0.9730$\pm$0.0007} & \underline{0.9400$\pm$0.0074} & \underline{0.9840$\pm$0.0033} \\ \hline
\textbf{P value of t-test} & \textbf{9.609e-07} & \textbf{0.0018956} & \textbf{0.157513} & \textbf{0.08213} \\ \hline
\end{tabular}
\end{table}    
To show that CODE-AE can generate transferable embedding through deconfounding uninteresting confounders while preserving true biological signals present in expression data even outside the in-vivo and in-vitro setting. We selected the gene expression data sets used in ADAE \cite{adae} to perform a similar evaluation process. Specifically, we chose the brain cancer expression data set with gender information as confounding factors and brain cancer subtype classification as target downstream tasks. We first performed encoder training with all unlabeled gene expression profiles regardless of gender. For ADAE \cite{adae} and CODE-AE, we selected the binary gender variable as the deconfounding target. After encoder training, we generated the latent embedding for all original gene expression profiles using different encoders. Then,  we built elastic net classifiers for cancer subtype prediction using the latent embedding of samples of one gender to predict the other gender samples. Following the evaluation procedure described in \cite{adae}, the classification performance measured in the area under the precision-recall curve (AUPRC) as well as area under the receiver operating curve (AUROC) of ten-fold cross-validation was reported in Table~3\vphantom{\ref{Tab:03}}. Besides, we performed a two-sample t-test on the average performance between CODE-AE and the best non-CODE-AE method in each setting, and its results are shown on the last row of Table~3\vphantom{\ref{Tab:03}}. We observed the same trends as those in the drug resistance prediction. Using the model built from female data to predict male data, CODE-AE-ADV significantly outperforms ADAE, the second-best performer measured by both AUROC and AUPRC. When applying the model trained from male data to predict female data, the performance of CODE-AE-ADV is slightly worse than CORAL, but the difference is not statistically significant. Both CODE-AE-ADV and CORAL significantly outperform the state-of-the-art deconfounding method ADAE (p-value \(\leq\) 0.05). Additionally, two other observations from the chemotherapy resistance experiments hold. Disentangling common and private features of different data modalities is essential for cell line to tissue transfer learning, and adversarial loss is more effective than MMD loss. 
\section{Conclusion}
In this paper, we introduce a new transfer learning framework CODE-AE to predict patient drug response from a supervised neural network model trained from cell line data. Extensive benchmark studies demonstrate the advantage of CODE-AE over the state-of-the-art in terms of both accuracy and robustness. The performance gain of CODE-AE mainly comes from (1) the unsupervised learning that combines unlabeled data from both cell lines and patient samples, (2) separation of shared common features cross cell lines and patient samples with unique embedding for cell lines or patients, and (3) adversarial training to optimize the similarity and difference between incoherent data sets. CODE-AE could be further improved in several directions. In contrast with cell line data from a pure population of cells, patient tissue data are mixtures of normal, abnormal, and infiltrated immune cells. We can further improve the CODE-AE by the deconvolution of patient gene expression data. We only use transcriptomics profiles to build the predictive model in this study. We can integrate additional omics data such as somatic mutations and copy number variants in the framework of cross-level information transmission \cite{he2020crosslevel}. Finally, we only apply CODE-AE to cancers. It will be interesting to test the performance of CODE-AE in other diseases besides cancers, which even do not have a large number of cell line data. In principle, CODE-AE can be applied to other transfer learning tasks with two data modalities with shared and unique features.  
\section*{Funding}
This work has been supported by the National Institute of General Medical Sciences of National Institute of Health (R01GM122845) and the National Institute on Aging of the National Institute of Health. (R01AD057555)\vspace*{-12pt}
\bibliographystyle{unsrt}
\bibliography{code-ae.bib}

\begin{thebibliography}{10}

\bibitem{danaee2017deep}
Padideh Danaee, Reza Ghaeini, and David~A Hendrix.
\newblock A deep learning approach for cancer detection and relevant gene
  identification.
\newblock In {\em Pacific symposium on biocomputing 2017}, pages 219--229.
  World Scientific, 2017.

\bibitem{gtex2017genetic}
GTEx Consortium et~al.
\newblock Genetic effects on gene expression across human tissues.
\newblock {\em Nature}, 550(7675):204, 2017.

\bibitem{goossens2015cancer}
Nicolas Goossens, Shigeki Nakagawa, Xiaochen Sun, and Yujin Hoshida.
\newblock Cancer biomarker discovery and validation.
\newblock {\em Translational cancer research}, 4(3):256, 2015.

\bibitem{CCLE}
Jordi Barretina, Giordano Caponigro, Nicolas Stransky, Kavitha Venkatesan,
  Adam~A Margolin, Sungjoon Kim, Christopher~J Wilson, Joseph Leh{\'a}r,
  Gregory~V Kryukov, Dmitriy Sonkin, et~al.
\newblock The cancer cell line encyclopedia enables predictive modelling of
  anticancer drug sensitivity.
\newblock {\em Nature}, 483(7391):603--607, 2012.

\bibitem{sakellaropoulos2019deep}
Theodore Sakellaropoulos, Konstantinos Vougas, Sonali Narang, Filippos Koinis,
  Athanassios Kotsinas, Alexander Polyzos, Tyler~J Moss, Sarina Piha-Paul, Hua
  Zhou, Eleni Kardala, et~al.
\newblock A deep learning framework for predicting response to therapy in
  cancer.
\newblock {\em Cell reports}, 29(11):3367--3373, 2019.

\bibitem{autoencoders}
Geoffrey~E Hinton and Richard~S Zemel.
\newblock Autoencoders, minimum description length, and helmholtz free energy.
\newblock {\em Advances in neural information processing systems}, 6:3--10,
  1994.

\bibitem{dae}
Pascal Vincent, Hugo Larochelle, Yoshua Bengio, and Pierre-Antoine Manzagol.
\newblock Extracting and composing robust features with denoising autoencoders.
\newblock In {\em Proceedings of the 25th international conference on Machine
  learning}, pages 1096--1103, 2008.

\bibitem{vae}
Diederik~P Kingma and Max Welling.
\newblock Auto-encoding variational bayes.
\newblock {\em arXiv preprint arXiv:1312.6114}, 2013.

\bibitem{ching2018opportunities}
Travers Ching, Daniel~S Himmelstein, Brett~K Beaulieu-Jones, Alexandr~A
  Kalinin, Brian~T Do, Gregory~P Way, Enrico Ferrero, Paul-Michael Agapow,
  Michael Zietz, Michael~M Hoffman, et~al.
\newblock Opportunities and obstacles for deep learning in biology and
  medicine.
\newblock {\em Journal of The Royal Society Interface}, 15(141):20170387, 2018.

\bibitem{weiss2016survey}
Karl Weiss, Taghi~M Khoshgoftaar, and DingDing Wang.
\newblock A survey of transfer learning.
\newblock {\em Journal of Big data}, 3(1):9, 2016.

\bibitem{DCC2014}
Eric Tzeng, Judy Hoffman, Ning Zhang, Kate Saenko, and Trevor Darrell.
\newblock Deep domain confusion: Maximizing for domain invariance.
\newblock {\em arXiv preprint arXiv:1412.3474}, 2014.

\bibitem{deepcoral2016}
Baochen Sun and Kate Saenko.
\newblock Deep coral: Correlation alignment for deep domain adaptation.
\newblock In {\em European conference on computer vision}, pages 443--450.
  Springer, 2016.

\bibitem{dann2016}
Yaroslav Ganin, Evgeniya Ustinova, Hana Ajakan, Pascal Germain, Hugo
  Larochelle, Fran{\c{c}}ois Laviolette, Mario Marchand, and Victor Lempitsky.
\newblock Domain-adversarial training of neural networks.
\newblock {\em The Journal of Machine Learning Research}, 17(1):2096--2030,
  2016.

\bibitem{adda2017}
Eric Tzeng, Judy Hoffman, Kate Saenko, and Trevor Darrell.
\newblock Adversarial discriminative domain adaptation.
\newblock In {\em Proceedings of the IEEE conference on computer vision and
  pattern recognition}, pages 7167--7176, 2017.

\bibitem{mDAE}
Minmin Chen, Zhixiang Xu, Kilian Weinberger, and Fei Sha.
\newblock Marginalized denoising autoencoders for domain adaptation.
\newblock {\em arXiv preprint arXiv:1206.4683}, 2012.

\bibitem{mtAE}
Muhammad Ghifary, W~Bastiaan~Kleijn, Mengjie Zhang, and David Balduzzi.
\newblock Domain generalization for object recognition with multi-task
  autoencoders.
\newblock In {\em Proceedings of the IEEE international conference on computer
  vision}, pages 2551--2559, 2015.

\bibitem{drcn}
Muhammad Ghifary, W~Bastiaan Kleijn, Mengjie Zhang, David Balduzzi, and Wen Li.
\newblock Deep reconstruction-classification networks for unsupervised domain
  adaptation.
\newblock In {\em European Conference on Computer Vision}, pages 597--613.
  Springer, 2016.

\bibitem{dsn}
Konstantinos Bousmalis, George Trigeorgis, Nathan Silberman, Dilip Krishnan,
  and Dumitru Erhan.
\newblock Domain separation networks.
\newblock In {\em Advances in neural information processing systems}, pages
  343--351, 2016.

\bibitem{MMD}
Arthur Gretton, Karsten~M Borgwardt, Malte~J Rasch, Bernhard Sch{\"o}lkopf, and
  Alexander Smola.
\newblock A kernel two-sample test.
\newblock {\em Journal of Machine Learning Research}, 13(Mar):723--773, 2012.

\bibitem{adae}
Ayse~B Dincer, Joseph~D Janizek, and Su-In Lee.
\newblock Adversarial deconfounding autoencoder for learning robust gene
  expression embeddings.
\newblock {\em bioRxiv}, 2020.

\bibitem{fols}
Mathieu Salzmann, Carl~Henrik Ek, Raquel Urtasun, and Trevor Darrell.
\newblock Factorized orthogonal latent spaces.
\newblock In {\em Proceedings of the Thirteenth International Conference on
  Artificial Intelligence and Statistics}, pages 701--708, 2010.

\bibitem{wgan}
Martin Arjovsky, Soumith Chintala, and L{\'e}on Bottou.
\newblock Wasserstein generative adversarial networks.
\newblock In {\em International conference on machine learning}, pages
  214--223. PMLR, 2017.

\bibitem{wgangp}
Ishaan Gulrajani, Faruk Ahmed, Martin Arjovsky, Vincent Dumoulin, and Aaron~C
  Courville.
\newblock Improved training of wasserstein gans.
\newblock In {\em Advances in neural information processing systems}, pages
  5767--5777, 2017.

\bibitem{CCLE2}
Mahmoud Ghandi, Franklin~W Huang, Judit Jan{\'e}-Valbuena, Gregory~V Kryukov,
  Christopher~C Lo, E~Robert McDonald, Jordi Barretina, Ellen~T Gelfand,
  Craig~M Bielski, Haoxin Li, et~al.
\newblock Next-generation characterization of the cancer cell line
  encyclopedia.
\newblock {\em Nature}, 569(7757):503--508, 2019.

\bibitem{GDSC1}
Wanjuan Yang, Jorge Soares, Patricia Greninger, Elena~J Edelman, Howard
  Lightfoot, Simon Forbes, Nidhi Bindal, Dave Beare, James~A Smith, I~Richard
  Thompson, et~al.
\newblock Genomics of drug sensitivity in cancer (gdsc): a resource for
  therapeutic biomarker discovery in cancer cells.
\newblock {\em Nucleic acids research}, 41(D1):D955--D961, 2012.

\bibitem{GDSC2}
Francesco Iorio, Theo~A Knijnenburg, Daniel~J Vis, Graham~R Bignell, Michael~P
  Menden, Michael Schubert, Nanne Aben, Emanuel Gon{\c{c}}alves, Syd Barthorpe,
  Howard Lightfoot, et~al.
\newblock A landscape of pharmacogenomic interactions in cancer.
\newblock {\em Cell}, 166(3):740--754, 2016.

\bibitem{xena}
Mary Goldman, Brian Craft, Angela Brooks, Jing Zhu, and David Haussler.
\newblock The ucsc xena platform for cancer genomics data visualization and
  interpretation.
\newblock {\em BioRxiv}, page 326470, 2018.

\bibitem{ben2019resistance}
Rotem Ben-Hamo, Alona Zilberberg, Helit Cohen, Keren Bahar-Shany, Chaim
  Wachtel, Jacob Korach, Sarit Aviel-Ronen, Iris Barshack, Danny Barash, Keren
  Levanon, et~al.
\newblock Resistance to paclitaxel is associated with a variant of the gene
  bcl2 in multiple tumor types.
\newblock {\em NPJ precision oncology}, 3(1):1--11, 2019.

\bibitem{tcga_dataset}
Evan~A Clayton, Toyya~A Pujol, John~F McDonald, and Peng Qiu.
\newblock Leveraging tcga gene expression data to build predictive models for
  cancer drug response.
\newblock {\em BMC bioinformatics}, 21(14):1--11, 2020.

\bibitem{tcga}
Carolyn Hutter and Jean~Claude Zenklusen.
\newblock The cancer genome atlas: creating lasting value beyond its data.
\newblock {\em Cell}, 173(2):283--285, 2018.

\bibitem{roc}
Tom Fawcett.
\newblock An introduction to roc analysis.
\newblock {\em Pattern recognition letters}, 27(8):861--874, 2006.

\bibitem{he2020crosslevel}
Di~He and Lei Xie.
\newblock A cross-level information transmission network for predicting
  phenotype from new genotype: Application to cancer precision medicine, 2020.

\end{thebibliography}

\end{document}